\documentclass{article}
\PassOptionsToPackage{numbers,sort&compress}{natbib}
\usepackage[preprint]{neurips_2026}
\workshoptitle{Managing Agents that Manage Agents: Workshop on Responsible Use of Meta-Agents}
\newif\ifcameraready
\camerareadyfalse
\usepackage[utf8]{inputenc}
\usepackage[T1]{fontenc}
\usepackage{hyperref}
\usepackage{url}
\usepackage{booktabs}
\usepackage{amsmath,amssymb,amsfonts}
\usepackage{nicefrac}
\usepackage{microtype}
\usepackage{xcolor}
\usepackage{graphicx}
\usepackage{tikz}
\usetikzlibrary{arrows.meta,positioning,calc}
\usepackage{array}
\usepackage{tabularx}
\usepackage{adjustbox}
\usepackage{enumitem}
\usepackage{multirow}
\usepackage{float}
\usepackage{pifont}

\hypersetup{
  hypertexnames=false,
  colorlinks=true,
  linkcolor=blue!45!black,
  citecolor=blue!45!black,
  urlcolor=blue!45!black,
  pdfcreator={pdfLaTeX}
}
\newcolumntype{Y}{>{\raggedright\arraybackslash}X}
\newcolumntype{C}[1]{>{\centering\arraybackslash}p{#1}}

\newcommand{\challenger}{\textsc{Challenger}}
\newcommand{\kgate}{\textsc{Knowledge Gate}}
\newcommand{\medbench}{\textsc{MedAgentsBench}}

\definecolor{softblue}{RGB}{235,243,250}
\definecolor{softorange}{RGB}{253,241,224}
\definecolor{softgreen}{RGB}{235,247,237}
\definecolor{softviolet}{RGB}{244,238,250}

\title{Challenge-Aware Knowledge Recruitment \\ with Meta-Agent Supervision \\ for Medical Multi-Agent Reasoning}

\author{%
  Riccardo Riccio \\
  University of Cambridge \\
  \And
  Haitz S\'aez de Oc\'ariz Borde \\
  University of Cambridge \\
  \And
  Pietro Li\`o \\
  University of Cambridge \\
}
\hypersetup{
  pdftitle={Challenge-Aware Knowledge Recruitment for Medical Multi-Agent Question Answering},
  pdfauthor={Anonymous Author(s)},
  pdfsubject={NeurIPS 2026 workshop paper}
}

\begin{document}
\maketitle
\begin{abstract}
Large language models perform strongly on standard medical QA but degrade on harder reasoning tasks, where multi-agent debate often fails to beat strong single-model prompting. We argue that a key limitation is that existing systems select experts and retrieve knowledge from the question alone, rather than from gaps that emerge during reasoning. We propose challenge-aware adaptive recruitment: an external CHALLENGER inspects the evolving reasoning trace and identifies specific factual or expertise gaps. These signals trigger targeted knowledge retrieval or mid-discussion specialist recruitment. On the MEDAGENTSBENCH hard subset (862 questions across nine biomedical datasets), upfront retrieval provides little benefit, while challenge-aware in-between retrieval performs better; on a multi-specialist backbone it improves accuracy on 8 of 9 datasets and adds 6 percentage points over Challenger Only. Ablations indicate benefits from both retrieval timing and gap-specific query signals, and a live web-search experiment shows that the timing advantage persists across retrieval backends. For specialist recruitment, self-triggered adaptation is unreliable, whereas Challenger-triggered recruitment raises GPT-4.1-mini MedQA accuracy from 46\% to 54\%, with positive direction on all tested generalization datasets. Naively combining retrieval and specialist recruitment degrades performance, highlighting the need to coordinate adaptive intervention channels. These results show that when and why external resources are recruited is a key design choice in multi-agent medical reasoning.

\end{abstract}

\section{Introduction}

In clinical practice, complex diagnoses often require expertise from multiple specialties. Hospitals therefore convene multidisciplinary teams (MDTs), bringing together specialist clinicians to discuss a case and agree on a diagnosis or treatment plan. By combining knowledge from several domains, MDTs can address problems that would be difficult for any individual clinician to solve alone.

However, an MDT is neither a fixed committee nor a group working from a predetermined body of evidence. The relevant specialties often become apparent only as the discussion unfolds. A radiological interpretation may reframe the clinical question, or an unexpected laboratory result may raise a possibility that had not previously been considered, prompting the team to seek an additional specialist opinion. Medical evidence is consulted in much the same way: not necessarily at the outset, but when a specific question becomes important to the team’s decision.

A similar idea has recently been explored in AI through multi-agent systems, in which multiple language-model agents with different roles or areas of expertise collaborate to solve a problem \citep{tang2024medagents,kim2024mdagents}. The agents receive medical role prompts, produce case analyses, discuss competing answers, and have their outputs aggregated into a final decision. These systems reproduce the structure of the panel, but not its adaptivity. On the difficult \medbench{} subset, published medical multi-agent systems do not consistently outperform strong prompting baselines \citep{tang2025medagentsbench}. More elaborate agent scaffolds are therefore not sufficient by themselves. Appendix~\ref{app:benchmark-context} gives the full benchmark comparison used to establish this setting.

A central design choice is when and how these systems use external medical knowledge. A common design retrieves evidence from the original question before the current trace has exposed a specific gap \citep{kim2024mdagents,wang2025medagent}. A fixed evidence brief cannot target a missing physiological fact, anatomical distinction, or literature claim that becomes apparent only during reasoning. It can also foreground a plausible but incorrect hypothesis before the panel has considered alternatives. We ask when evidence should be introduced and which reasoning gap should determine the query.

\begin{figure}[t]
  \centering
  \begin{tikzpicture}[
  font=\scriptsize,
  >=Latex,
  box/.style={draw, rounded corners=2pt, align=center, minimum height=11mm, inner sep=3pt},
  fixed/.style={box, fill=black!4, draw=black!55, text width=26mm},
  worker/.style={box, fill=blue!10, draw=blue!55!black, text width=25mm},
  supervisor/.style={box, fill=orange!18, draw=orange!70!black, text width=23mm},
  knowledge/.style={box, fill=green!12, draw=green!55!black, text width=26mm},
  expert/.style={box, fill=violet!12, draw=violet!65!black, text width=26mm},
  control/.style={box, fill=black!6, draw=black!60, text width=23mm},
  panlab/.style={font=\scriptsize\bfseries, anchor=west},
  arrow/.style={->, line width=0.55pt},
  advisory/.style={->, line width=0.55pt, dashed}
]

  \node[panlab] at (-1.3,1.95) {(a) Conventional design};

  \node[knowledge] (ev1)   at (0,0.95)   {Evidence retrieved\\from the\\question text};
  \node[expert]    (ex1)   at (0,-0.95)  {Expert team selected\\from the\\question text};
  \node[fixed]     (fix1)  at (4.0,0)    {Fixed expert team\\and evidence,\\set before reasoning};
  \node[worker]    (disc1) at (8.0,0)    {Multi-round\\discussion};
  \node[control]   (ans1)  at (12.0,0)   {Final answer\\majority vote};

  \draw[arrow] (ev1) -- (fix1);
  \draw[arrow] (ex1) -- (fix1);
  \draw[arrow] (fix1) -- (disc1);
  \draw[arrow] (disc1) -- (ans1);

  \node[panlab] at (-1.3,-2.55) {(b) Challenge-aware adaptive recruitment (ours)};

  \node[worker]     (back)  at (0,-4.40)      {Initial reasoning\\backbone\\generalist or\\specialist team};
  \node[supervisor] (chal)  at (2.90,-4.40)   {Supervisory\\Challenger\\identify gap $\delta_t$};
  \node[knowledge]  (kn)    at (6.05,-3.50)   {Evidence channel\\Gate $\rightarrow$ retrieval\\$\rightarrow$ brief $B_t$};
  \node[expert]     (ex2)   at (6.05,-5.30)   {Specialist channel\\re-evaluation\\$\rightarrow$ recruited agents};
  \node[worker]     (disc2) at (9.10,-4.40)   {Multi-round\\discussion with the\\added resources};
  \node[control]    (ans2)  at (12.00,-4.40)  {Final answer\\majority vote or\\reconsideration};

  \draw[arrow] (back) -- (chal);
  \draw[arrow] (chal) -- node[above left, pos=.62]{factual gap} (kn);
  \draw[arrow] (chal) -- node[below left, pos=.62]{expertise gap} (ex2);
  \draw[arrow] (kn) -- (disc2);
  \draw[arrow] (ex2) -- (disc2);
  \draw[arrow] (disc2) -- (ans2);
  \draw[advisory] (disc2.south) |- ++(0,-13mm) -| node[below, pos=.36]{up to three rounds} (back.south);
\end{tikzpicture}
  \caption{Challenge-aware supervision in medical multi-agent QA\@.
    \textbf{(a)} Conventional designs fix the expert team and potentially retrieve evidence
    from the question text before any reasoning has taken place, then discuss over multiple
    rounds and produce a final answer.
    \textbf{(b)} In the proposed framework the initial backbone reasons first and receives
    an external Challenger critique between rounds, after which support is recruited
    conditioned on the gap identified. The primary medical-evidence channel retrieves a
    curated brief, while the complementary specialist channel can add another
    role-conditioned agent. The active backbone then continues reasoning before final
    aggregation.}

  \label{fig:framework}
\end{figure}
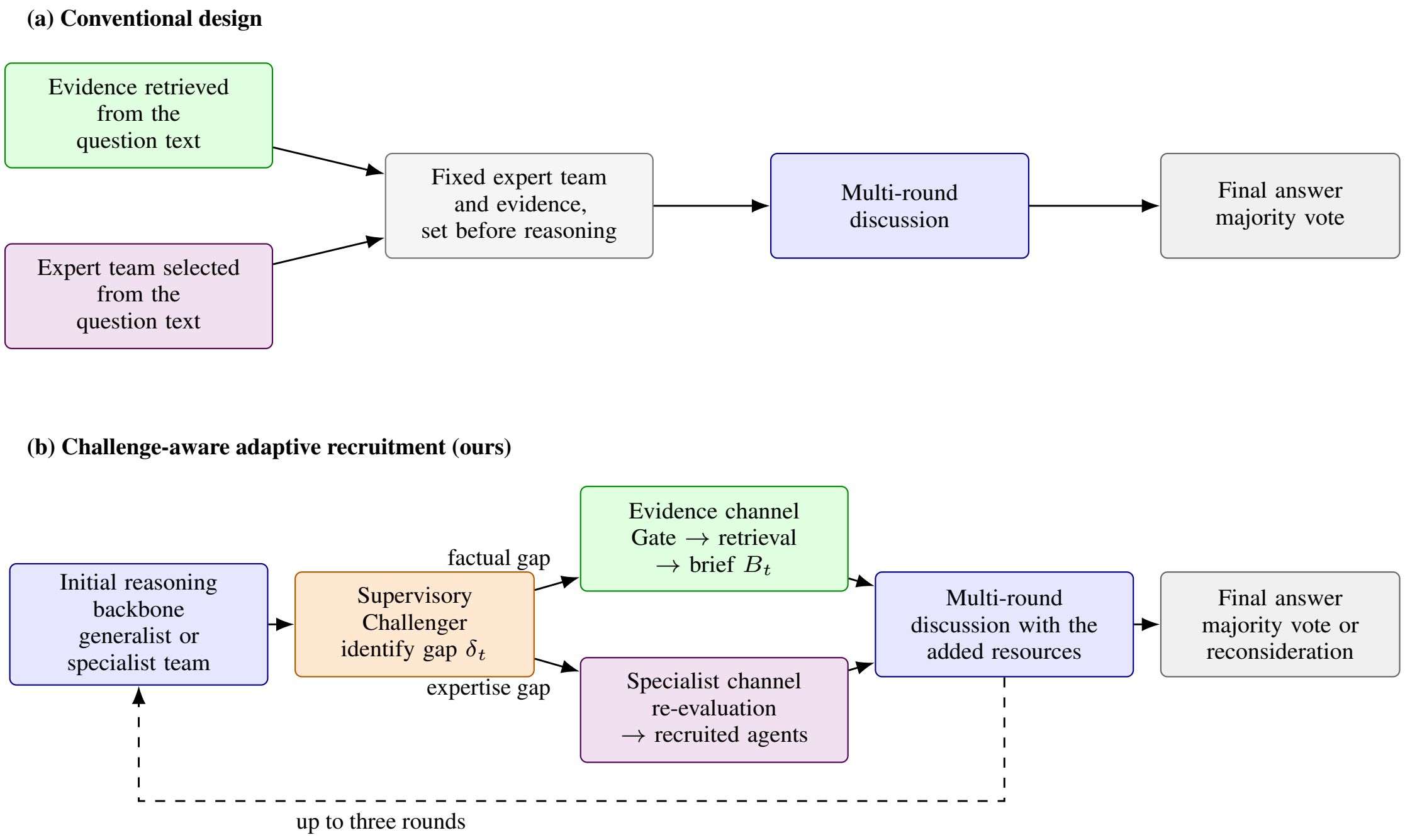

We introduce challenge-aware knowledge recruitment for this setting. An external \challenger{} (the meta-agent) sits outside the reasoning panel and reviews the evolving clinical trace. It identifies a concrete weakness, such as an unresolved differential diagnosis or an unsupported physiological claim, without supplying the missing content. A separate \kgate{} decides whether the diagnosed gap is retrievable, formulates a narrow query, and inserts a curated evidence brief between reasoning rounds. Diagnosis and retrieval are handled by different components. This makes the behavior of the system closer to the way an MDT operates in practice, where both the composition of the team and the evidence consulted are decided during the discussion rather than fixed before it. Figure~\ref{fig:framework} summarizes the design.

We additionally use the same supervisory diagnosis to study specialist recruitment. Existing dynamic medical-agent systems can ask the current agents whether another specialty is needed \citep{wu2025knowledge}, but this assumes that a confidently wrong panel can recognize its own missing perspective. We compare that self-triggered mechanism with recruitment after the Challenger has made a specific gap explicit. A final experiment studies how the evidence and specialist channels interact when both are active.

The experiments use medical multiple-choice QA as a controlled setting for adaptive supervision. The \medbench{} hard subset retains questions on which fewer than half of the evaluated models answer correctly. It spans clinical reasoning, biomedical literature interpretation, licensing examinations, scientific knowledge, and specialist-level medical tasks \citep{tang2025medagentsbench,zuo2025medxpertqa,wang2024mmlu}. While this benchmark does not fully reflect how decisions are made in a real hospital, it allows the effects of adaptive mechanisms in multi-agent systems to be studied systematically, and the findings can inform the design of future medical multi-agent decision-making systems.

The evaluation is organized around four questions. \textbf{RQ1} asks whether gap-targeted in-between medical evidence retrieval outperforms static upfront retrieval. \textbf{RQ2} asks whether the result holds with another retrieval backend and on a multi-specialist medical pipeline. As a complementary analysis, \textbf{RQ3} asks whether medical agents can identify missing expertise and whether an external Challenger improves the recruitment decision. \textbf{RQ4} then studies whether specialist and knowledge recruitment can be combined without interference. The contributions are:
\begin{itemize}[leftmargin=1.45em,itemsep=1.5pt,topsep=2pt]
  \item a staged medical retrieval study that evaluates evidence timing and access to the Challenger-derived query signal, together with a live web-search validation;
  \item evidence that gap-targeted knowledge recruitment transfers to a multi-specialist backbone, multiple base models, and a distinct retrieval backend;
  \item a complementary specialist-recruitment study that diagnoses failures in self-assessed expertise needs and isolates the effects of external challenge and adaptive recruitment; and
  \item an analysis of interference between the two intervention channels, supported by full cross-model results and three medical case studies in the appendix.
\end{itemize}
The appendix also reports the complete benchmark baselines, the KAMAC implementation check, full per-dataset results, transfer runs, run-to-run variance, and detailed experimental settings.

\section{Related work}

\paragraph{Retrieval during reasoning.}
Retrieval-augmented generation provides language models with external evidence rather than relying only on parametric memory \citep{lewis2020retrieval}. Medical RAG systems combine biomedical corpora with specialized retrievers \citep{xiong2024benchmarking}; iMedRAG refines an initial medical query through targeted follow-up questions \citep{xiong2024improving}. Several medical reasoning systems and agentic pipelines incorporate retrieved knowledge \citep{kim2024mdagents,lu2024clinicalrag,wang2025medagent}. A common design obtains evidence from the original input before the current reasoning trace has exposed a specific gap. Reasoning can also be sensitive to the order in which contextual premises are presented \citep{chen2024premise}. Tool-MAD and debate-augmented RAG revisit retrieval during discussion \citep{jeong2026tool,hu2025removal}. The systems reviewed here do not jointly evaluate retrieval stage and access to a separately diagnosed gap. Our staged four-condition ablation evaluates both in medical QA\@.

\paragraph{Multi-agent reasoning and aggregation.}
Multi-agent debate asks several language-model agents to propose, critique, and revise solutions before aggregation \citep{du2023improving}. Medical variants assign role-conditioned specialties and run structured discussion rounds \citep{tang2024medagents}, while complexity-calibrated systems route medical questions to solo, small-group, or larger-group configurations \citep{kim2024mdagents}. ReConcile uses confidence-weighted round-table discussion \citep{chen2024reconcile}, and decision-rule studies show that voting and consensus choices materially affect outcomes \citep{kaesberg2025voting}. More generally, additional inference-time reasoning and sampling can improve language-model performance through chain-of-thought and self-consistency \citep{wei2022chain,wang2022self}. On difficult medical subsets, however, extra multi-agent discussion does not guarantee an advantage over those strong single-model strategies \citep{tang2025medagentsbench}. We focus on how a supervisory component routes additional computation to evidence or, secondarily, another role-conditioned perspective.

\paragraph{Dynamic teams and external critique.}
KAMAC permits a medical-agent team to expand when agents self-assess that relevant knowledge is missing \citep{wu2025knowledge}. This makes team composition adaptive, but places the trigger inside the same medical reasoning process that may be confidently wrong. Reflexion incorporates verbal feedback across attempts \citep{shinn2023reflexion}, while devil's-advocate and Catfish-style agents introduce structured dissent to counter premature agreement \citep{kim2024debate,wang2025silence}. These systems show that external criticism can alter deliberation. Here the critique is used first to guide targeted evidence retrieval and, in a complementary study, to inform whether a medical panel should recruit another role-conditioned specialist; the Challenger does not itself supply the missing clinical content.

\section{Challenge-aware medical multi-agent framework}
\label{sec:framework}

\subsection{Supervising an evolving medical reasoning trace}

Let $q$ be a medical multiple-choice question with answer options $O$, $A_t$ the active reasoning backbone at turn $t$, represented by one generalist agent or a panel of role-conditioned medical specialists, and $h_t$ the accumulated reasoning state. The active backbone first produces an initial reasoning trace rather than immediately invoking external evidence or another specialist. An independent Challenger evaluates that trace,
\begin{equation}
  C(q,O,h_t)=(j_t,\delta_t), \qquad j_t\in\{0,1\},
  \label{eq:challenger}
\end{equation}
where $j_t=1$ denotes a material weakness and $\delta_t$ contains a structured characterization of the gap and a targeted challenge. The structured output covers issues such as a logical inconsistency, a missing differential diagnosis, insufficient medical evidence for an answer, or an unresolved conflict among options. The critique is appended to the reasoning history before any downstream intervention decision. Throughout the paper, ``expert'' and ``specialist'' refer to role-conditioned model instances. The Challenger identifies a reasoning need; it does not directly provide the answer or the missing content.

As shown in Figure~\ref{fig:framework}, the primary evidence channel and the complementary specialist channel act after an initial reasoning step and before further deliberation or reconsideration, so their input reflects the clinical or factual uncertainty exposed by the current trace.

\subsection{Adaptive intervention channels}

\paragraph{Primary knowledge-recruitment channel.}
A separate \kgate{} evaluates whether the current medical reasoning gap is addressable through retrieval. Given $(q,O,h_t,\delta_t)$, it either returns no query or constructs a narrow query $q_t^{\mathrm{ret}}$ targeting the unresolved clinical or biomedical point. iMedRAG retrieves $k=16$ passages from PubMed abstracts and medical textbooks, and an Evidence Curator filters and compresses them into a brief $B_t$ containing five to eight evidence points before reconsideration. The gate is independent of the Challenger: the critique is advisory context, while the gate makes its own decision about whether external medical evidence is likely to help. Thus a critique does not automatically trigger retrieval.

\paragraph{Complementary specialist-recruitment channel.}
The multi-specialist backbone begins with two specialist-role agents selected from the question. After each turn, active agents assess whether the Challenger's identified gap requires a medical perspective not represented in $A_t$. If at least one agent flags a missing specialty, a recruiter proposes non-duplicate role labels and adds at most two agents in that turn, subject to a maximum panel size of five. Recruited agents receive the reasoning trace up to the intervention and participate in subsequent discussion. This produces two triggers for the same recruitment mechanism: \emph{self-triggered} recruitment asks the panel before an external gap is available, while \emph{Challenger-informed} recruitment asks after $\delta_t$ has made a specific medical reasoning weakness explicit.

\paragraph{Aggregation and channel coordination.}
The multi-specialist medical backbone runs three rounds, with a moderator selecting the final option by majority vote over active agents. In the naive combined system, recruited specialists receive the same evidence brief as the original panel. This can make the two channels redundant or produce correlated errors. We therefore also evaluate an independent variant in which recruited specialists do not receive $B_t$ directly and supply auxiliary assessments to the moderator rather than reshaping the shared evidence-informed discussion. This tests whether information-flow separation improves the combination within this medical QA pipeline.

\subsection{Ablation conditions}

The knowledge study compares No Retrieval, Upfront Retrieval, In-Between Blind Retrieval, and In-Between Challenger-Aware Retrieval. For the complementary specialist study, we denote Challenger activation by $c\in\{0,1\}$ and specialist recruitment by $r\in\{0,1\}$: Fixed Panel ($c_0r_0$), Self-Triggered Recruitment ($c_0r_1$), Challenger Only ($c_1r_0$), and Challenger + Recruitment ($c_1r_1$). Knowledge transfer to the multi-specialist backbone is $c_1r_0k_1$; the naive combined system is $c_1r_1k_1$; and the independent coordination variant is $c_1r_1k_1^{\mathrm{ind}}$. Descriptive labels are used in the main tables, with full condition definitions in Appendix~\ref{app:implementation} and Table~\ref{tab:conditions-app}.

\section{Experimental setup}
\label{sec:setup}

\paragraph{Benchmark and models.}
All primary experiments use the 862-question \medbench{} hard subset, constructed by retaining questions answered correctly by fewer than 50\% of evaluated models \citep{tang2025medagentsbench}. It contains nine sub-datasets: MedQA, PubMedQA, MedMCQA, MedBullets, MMLU, MMLU-Pro, MedExQA, MedXpertQA-R, and MedXpertQA-U. GPT-4o-mini is used for full benchmark comparisons because it matches the model setting used by published benchmark comparisons and makes broad ablation feasible. GPT-4.1-mini tests transfer to a stronger model. All seven principal multi-specialist conditions are also evaluated with grok-4-1-fast-non-reasoning in Appendix~\ref{app:grok}. The self-assessment diagnostic uses the repository's 894-question hard set, which includes 32 additional repository questions beyond the primary 862. Appendix~\ref{app:benchmark-context} gives the complete baseline table and the KAMAC implementation check.

\paragraph{Protocol and implementation.}
The primary metric is pass@1 accuracy. Temperature is zero and the API seed is 42. API-level determinism is not guaranteed, so differences around one percentage point are treated cautiously. Following the \medbench{} evaluation protocol, the main tables report one pass@1 run per condition. Appendix~\ref{app:variance} reports three-run means and sample standard deviations for the GPT-4o-mini conditions. The system is implemented in Python without an external orchestration library. Agents share OpenAI-format message histories; recruited agents receive the prior trajectory; the Challenger, Knowledge Gate, recruiter, and moderator emit bracketed structured fields; and answer extraction is constrained to each dataset's valid option set. Questions are parallelized across four Optuna workers with per-run SQLite state \citep{akiba2019optuna}, while turns within a question execute sequentially. Model calls use an Azure OpenAI endpoint. The live-search experiment uses GPT-4o-mini in a separate complexity-calibrated reasoning backbone and replaces iMedRAG with a Claude Haiku model equipped with web search. Search cost limits this experiment to 100 MedQA questions.

\paragraph{Evaluation logic.}
RQ1 compares retrieval stage among the no-retrieval, upfront, and blind in-between conditions, then tests access to the Challenger diagnosis at fixed in-between timing. RQ2 tests the same design principle with live search, transfers it to the multi-specialist backbone, and checks the qualitative trend on a different model family. RQ3 combines a question-level self-assessment diagnostic with a $2\times2$ specialist-recruitment ablation. RQ4 compares knowledge-only intervention against naive and independent combinations in the same medical pipeline. Appendix~\ref{app:benchmark-context} reports benchmark context and implementation validation. Appendix~\ref{app:fulltables} contains full per-dataset tables, Appendix~\ref{app:grok} contains all Grok conditions, and Appendix~\ref{app:variance} reports run-to-run variance. Appendix~\ref{app:cases} and Appendix~\ref{app:medical-discussion} provide three medical case studies and their cross-case interpretation.

\section{Results}
\label{sec:results}

\subsection{RQ1: Medical evidence retrieval benefits from retrieval stage and a gap-specific query signal}

Table~\ref{tab:retrieval-main} compares retrieval timing and query signal on the single-generalist backbone. Conventional upfront iMedRAG adds only 0.3 percentage points over the no-retrieval baseline (23.7\% versus 23.4\%) and reduces accuracy on MedQA and PubMedQA\@. Delaying retrieval until after the initial answer, while keeping the gate blind to the Challenger, reaches 24.4\%. Giving the in-between gate access to the diagnosed gap reaches 26.3\%. The overall ordering
\[
\text{No retrieval} < \text{Upfront} < \text{In-between blind} < \text{In-between Challenger-aware}
\]
is consistent with contributions from both retrieval stage and the Challenger-derived query signal. The intermediate differences are small, while the wider endpoint gap and the repeated direction across backbones provide the stronger evidence. Full per-dataset values appear in Appendix Table~\ref{tab:knowledge-full-app}.

\begin{table}[t]
\caption{Knowledge-recruitment ablation. The corpus study uses GPT-4o-mini on all 862 questions. The live-search study uses a distinct complexity-calibrated backbone on 100 MedQA questions. In both retrieval backends, the best result introduces evidence after an exposed reasoning gap. N/A denotes not applicable; N/E denotes not evaluated.}
\label{tab:retrieval-main}
\centering
\small
\begin{tabular}{lllrr}
\toprule
Condition & Timing & Query signal & Corpus overall & Web MedQA\\
\midrule
No retrieval & N/A & N/A & 23.4 & 36.0\\
Upfront retrieval & Before reasoning & Question text & 23.7 & 41.0\\
In-between blind & After initial trace & Agent state & 24.4 & N/E\\
Challenger-aware & After initial trace & Diagnosed gap & \textbf{26.3} & \textbf{47.0}\\
\bottomrule
\end{tabular}
\end{table}

\subsection{RQ2: The knowledge-recruitment benefit transfers across pipelines and retrieval backends}

The live web-search experiment tests whether the advantage is an artifact of the fixed iMedRAG corpus. Using the same search backend in both conditions, upfront retrieval scores 41\% and in-between Challenger-aware retrieval scores 47\% on MedQA, compared with 36\% without retrieval (Table~\ref{tab:retrieval-main}). Since both retrieval conditions use the same source, the six-percentage-point difference is not explained by corpus choice alone and is consistent with routing retrieval after a diagnosed gap. This experiment uses one 100-question subset and one search agent.

Knowledge recruitment also transfers to the richer multi-specialist backbone. Adding Challenger-aware retrieval to Challenger Only ($c_1r_0k_1$ versus $c_1r_0$) improves GPT-4o-mini accuracy from 20.3\% to 26.3\%, with gains on eight of nine datasets. This 6.0-percentage-point improvement is the strongest single intervention in the main study. GPT-4.1-mini reaches 36.0\% under the same knowledge-only configuration. The full multi-specialist results are in Appendix Tables~\ref{tab:combined-4o-app} and~\ref{tab:combined-41-app}; the qualitative trend also transfers to Grok (Appendix~\ref{app:grok}), and the three-run GPT-4o-mini analysis preserves the ordering (Appendix~\ref{app:variance}).

\subsection{RQ3: Complementary analysis of specialist recruitment}

Panel A of Table~\ref{tab:expert-main} reports the controlled $2\times2$ ablation. On GPT-4o-mini, the Challenger improves the fixed panel from 18.1\% to 20.3\%, and adding recruitment reaches 21.1\%. The 0.8-percentage-point increment over Challenger Only is small in aggregate and varies across datasets: recruitment adds 8 percentage points on PubMedQA and 7 percentage points on MedExQA, while MedQA and MMLU decline. The focused GPT-4.1-mini result is larger. On 100 MedQA questions, Challenger + Recruitment reaches 54\%, compared with 46\% for Challenger Only, 44\% for the fixed panel, and 43\% for self-triggered recruitment. The recruitment rate also rises from 34\% under self-triggering to 41\% after the Challenger exposes a gap. Appendix Table~\ref{tab:expert-medqa-app} gives the complete MedQA ablation, including recruitment rates. The gain over Challenger Only remains positive on the three additional GPT-4.1-mini subsets tested: 4 percentage points on PubMedQA, 2 percentage points on MedMCQA, and 7 percentage points on MMLU-Pro (Appendix Table~\ref{tab:expert-gen-app}).

Panel B of Table~\ref{tab:expert-main} provides the supporting self-assessment diagnostic. GPT-4.1-mini is more accurate than GPT-4o-mini (24.7\% versus 17.3\%), but requests another agent on only 3 of 894 questions, compared with 219 for GPT-4o-mini. Despite a 75.3\% error rate, it recruits on 0.3\% of questions. The raw counts show how sharply recruitment behavior differs between the two models. With only two models, the central observation is model dependence rather than a general trend with capability. Self-declared specialist need can become nearly inactive despite frequent errors. The full GPT-4o-mini specialist table appears in Appendix Table~\ref{tab:expert-full-app}.

\begin{table}[t]
\caption{Complementary specialist-recruitment study. \textbf{A}: pass@1 accuracy (\%) for the controlled $2\times2$ ablation. \textbf{B}: question-level self-assessment behavior on the 894-question hard set. Challenger-informed recruitment reverses the direction of the self-triggered mechanism; the strongest effect appears on GPT-4.1-mini MedQA\@.}
\label{tab:expert-main}
\centering
\small
\textbf{A. Controlled specialist ablation}\\[-1mm]
\begin{tabular}{lrrrr}
\toprule
Setting & Fixed panel & Self-recruit & Challenger & Challenger + recruit\\
\midrule
GPT-4o-mini, 862 questions & 18.1 & 16.9 & 20.3 & \textbf{21.1}\\
GPT-4.1-mini, MedQA-100    & 44.0 & 43.0 & 46.0 & \textbf{54.0}\\
\bottomrule
\end{tabular}

\vspace{2mm}
\textbf{B. Self-assessment diagnostic}\\[-1mm]
\begin{tabular}{lrrrr}
\toprule
Model & Accuracy & Error & Recruited & Mean agents\\
\midrule
GPT-4o-mini  & 17.3 & 82.7 & 219/894 (24.5\%) & 2.09\\
GPT-4.1-mini & 24.7 & 75.3 &   3/894 (0.3\%)  & 1.01\\
\bottomrule
\end{tabular}
\end{table}

\subsection{RQ4: Interaction between specialist and evidence channels}

We next ask whether the complementary specialist channel adds to the stronger evidence channel. Activating both does not yield additive gains. As Table~\ref{tab:coordination-main} shows, the naive combination reduces GPT-4o-mini from 26.3\% to 23.0\% and GPT-4.1-mini from 36.0\% to 33.9\%. The regression appears on eight of nine GPT-4o-mini datasets. Both channels respond to the same Challenger diagnosis: the evidence brief and the recruited specialist may address the same weakness, and the specialist enters with a framing already established by retrieved evidence. Majority voting then treats correlated contributions as independent votes.

The independent redesign withholds the evidence brief from recruited specialists and passes their assessments to the moderator as auxiliary inputs. It improves GPT-4.1-mini from 36.0\% to 37.0\%, but remains below knowledge-only on GPT-4o-mini (23.9\% versus 26.3\%). Grok also shows mixed combined results. Neither combined design improves consistently across models. In this medical QA architecture, adding a specialist channel to an effective evidence channel can reduce accuracy unless the triggers, information flows, and aggregation roles are coordinated. Full per-dataset comparisons appear in Appendix Tables~\ref{tab:combined-4o-app} and~\ref{tab:combined-41-app}.

\begin{table}[t]
\caption{Coordination of specialist and knowledge interventions on the 862-question benchmark. Parentheses give the change from Challenger-aware knowledge recruitment alone. The independence-oriented redesign recovers the loss only for GPT-4.1-mini.}
\label{tab:coordination-main}
\centering
\small
\begin{tabular}{lrrr}
\toprule
Model & Knowledge only & Naive specialist + knowledge & Independent specialist + knowledge\\
\midrule
GPT-4o-mini  & 26.3 & 23.0 ($-3.3$) & 23.9 ($-2.4$)\\
GPT-4.1-mini & 36.0 & 33.9 ($-2.1$) & \textbf{37.0} ($+1.0$)\\
\bottomrule
\end{tabular}
\end{table}

\section{Medical case analysis and discussion}
\label{sec:discussion}

\paragraph{When knowledge recruitment helps.}
The strongest result concerns evidence routing. Upfront retrieval uses the same corpus as in-between retrieval but is less effective. Gap-targeted retrieval also transfers to a multi-specialist backbone and a live-search backend. In these pipelines, the current reasoning trace supplies information that the original question does not: it reveals which fact, comparison, or mechanism has become decisive.

\paragraph{Relevant evidence can still target the wrong distinction.}
An upfront query is organized around the surface form of the question. It may retrieve accurate material while emphasizing the wrong diagnosis or anatomical structure. In-between retrieval lets the agents form an initial interpretation first, then uses the exposed disagreement to construct a narrower query. Table~\ref{tab:case-main} illustrates this pattern for MedQA question medqa-477. The upfront evidence correctly describes the radial nerve but reinforces the wrong anatomical target. The gap-specific run retrieves evidence about the musculocutaneous nerve and changes the final answer. Appendix~\ref{app:cases} presents this case in full together with two additional medical traces.

\begin{table}[t]
\caption{Compact trace comparison for medqa-477. The question asks for the motor weakness associated with lateral-forearm sensory loss after heavy lifting; the correct answer is elbow-flexion weakness (A). The two retrieval configurations reinforce different anatomical hypotheses in this trace.}
\label{tab:case-main}
\centering
\small
\begin{tabularx}{\linewidth}{>{\raggedright\arraybackslash}p{23mm}YY}
\toprule
Stage & Upfront retrieval & Challenger-aware in-between retrieval\\
\midrule
Initial framing & Retrieved material centers the radial nerve and wrist extension. & Workers reason without injected evidence; the Challenger questions the radial-nerve interpretation.\\
Supervisory action & Later criticism mentions alternatives but does not dislodge the established frame. & The Knowledge Gate targets lateral-forearm innervation and musculocutaneous involvement.\\
Evidence effect & Accurate radial-nerve facts stabilize the wrong target. & The brief links musculocutaneous injury to elbow-flexion weakness and lateral-forearm sensory loss.\\
Final answer & (D), wrist extension (incorrect). & (A), elbow flexion (correct).\\
\bottomrule
\end{tabularx}
\end{table}

\paragraph{Adaptive allocation of inference-time computation.}
The framework allocates additional inference-time computation when the current trace exposes a gap. Those calls are assigned to different operations: trajectory evaluation, retrieval gating, evidence curation, and answer revision. The complementary specialist condition adds another role-conditioned analysis. This is different from asking one agent to repeat the same reasoning loop. Chain-of-thought and self-consistency already show that test-time computation can improve language-model performance \citep{wei2022chain,wang2022self}. The present ablations suggest that the destination of that computation matters as well. Upfront retrieval, self-triggered recruitment, and naive joint activation all use extra calls, yet each underperforms the best knowledge-only condition.

\paragraph{Why the Challenger can assist specialist recruitment.}
An agent asked whether its own panel lacks expertise must evaluate a reasoning process in which it has already participated. The Challenger receives a separate evaluative prompt and can reopen a line of reasoning after the panel has converged. Its critique turns generic uncertainty into a more specific task, such as asking for an electrolyte-management perspective. The MedQA-34 case in Appendix~\ref{app:cases} shows how that change in framing affects the recruited role and the final vote.

\paragraph{The two intervention channels require coordination.}
A recruited specialist and an evidence brief can address the same diagnosed gap. If both enter the shared discussion with the same framing, their contributions become correlated and majority voting can count redundant evidence more than once. A future controller could classify the exposed problem as a factual gap, a missing specialty, or a reasoning conflict, then route to one intervention or assign distinct roles to both. It may also need to preserve independent information paths and record which agents received the evidence brief. Appendix~\ref{app:medical-discussion} compares these mechanisms across the three qualitative cases.

\section{Limitations and responsible use}
\label{sec:responsible}

\paragraph{Limitations.}
The main results follow the benchmark protocol and report one pass@1 run per condition. Three-run GPT-4o-mini experiments keep total-accuracy standard deviation below one percentage point and preserve the main directional comparisons, although the +0.8 and +1.0 percentage-point effects remain small. The main text evaluates two OpenAI models, with Grok used as an additional transfer model. Open-source models, larger systems, and a wider range of providers remain untested. The Challenger, Knowledge Gate, Evidence Curator, recruiter, and moderator also depend on prompt design, and we do not include a prompt-sensitivity study.

The architecture intentionally uses more inference-time calls than its simplest baselines. We do not include a call-matched control, so the experiments do not fully separate the benefit of targeted routing from the benefit of a larger computation budget. The ablations nevertheless show that the extra calls are not interchangeable: upfront retrieval, self-triggered recruitment, and naive joint activation all perform worse than Challenger-aware knowledge recruitment. Cost and latency are not optimized, and the coordination problem between the two intervention channels remains open.

\paragraph{Responsible-use statement.}
The evaluation covers multiple-choice medical questions, not patient care, clinical workflows, or treatment planning. No clinician evaluated the Challenger critiques, generated specialist roles, retrieved evidence, or final answers. The system must therefore not be used as autonomous clinical decision support. Its ``specialists'' are role-conditioned language models, and an incorrect answer can acquire false authority when several agents agree. Retrieved material may also be outdated, incomplete, biased, or irrelevant to the specific patient population.

Any future clinical study would require qualified human oversight and independent verification against authoritative sources. The system should log the Challenger critique, recruitment decision, retrieval query, evidence provenance, generated role, and final aggregation so that failures can be audited. Retrieval sources should be screened for recency, conflicts of interest, demographic bias, and unsupported claims. Generated specialties should be monitored for stereotypes and misleading claims of expertise. Rate limits, halt controls, cost budgets, and latency thresholds are also needed because adaptive recruitment can expand during execution. Evaluation should include adversarial retrieval, low-quality role agents, disagreement with clinicians, uncertainty calibration, and abstention before the system is considered for any real clinical workflow.

\section{Conclusion}

We introduced challenge-aware knowledge recruitment with meta-agent supervision for medical multi-agent reasoning. A supervisory Challenger identifies a local gap in the evolving reasoning trace, and a Knowledge Gate directs additional inference-time computation to targeted retrieval and evidence curation. The method reaches 26.3\%, compared with 23.7\% for upfront retrieval and 23.4\% without retrieval. On a multi-specialist backbone, it adds 6.0 percentage points over Challenger Only, and the same preference for post-trace, gap-targeted retrieval appears in the live-search study. The complementary specialist analysis shows that meta-level external supervision can turn an unreliable self-triggered mechanism into a useful recruitment strategy. Naively activating both intervention channels reduces accuracy. These results support using supervisory signals from the evolving reasoning process to adaptively assign additional knowledge and agents, with explicit coordination when several intervention channels are available.

\label{lastmainpage}
\clearpage

{\small
\bibliographystyle{plainnat}
\bibliography{references}
}

\clearpage
\appendix
\section{Experimental conditions and implementation details}
\label{app:implementation}

\subsection{Backbones and condition notation}

Table~\ref{tab:pipelines-app} summarizes the three reasoning backbones, and Table~\ref{tab:conditions-app} defines every ablation condition. The single-generalist backbone isolates medical-evidence timing and query signal without multi-agent interaction. The multi-specialist backbone begins with two question-conditioned medical role agents and uses three rounds of three turns, with the Challenger evaluating the reasoning after each turn. A turn is one cycle in which every active agent responds; a round contains three such cycles. Two agents are the smallest panel that permits disagreement without making a large fixed team the baseline. The same three-round, three-turn budget is used across all multi-specialist conditions. A third complexity-calibrated backbone is used only for the live web-search validation. It classifies questions as Basic, Intermediate, or Advanced, calibrates the depth of the critique, and routes them to a solo, small-group, or multi-group discussion configuration.

\begin{table}[H]
\caption{Pipeline configurations used across the experiments.}
\label{tab:pipelines-app}
\centering
\small
\begin{tabularx}{\linewidth}{lYYY}
\toprule
Configuration & Initial backbone & Discussion structure & Purpose\\
\midrule
Single-generalist & One generalist agent & Initial answer, challenge, optional retrieval, reconsideration & Retrieval timing and query-signal ablation\\
Multi-specialist & Two question-conditioned specialist roles & Three rounds with three turns per round & Specialist recruitment, knowledge transfer, and channel coordination\\
Complexity-calibrated & Basic, Intermediate, or Advanced route & Solo, small-group, or multi-group discussion & Live web-search validation\\
\bottomrule
\end{tabularx}
\end{table}

\begin{table}[H]
\caption{Complete ablation-condition reference. $c$, $r$, and $k$ denote Challenger, specialist recruitment, and knowledge recruitment; ``ind'' denotes the independent coordination redesign.}
\label{tab:conditions-app}
\centering
\small
\begin{tabularx}{\linewidth}{lYYY}
\toprule
Condition & Active components & Backbone / purpose & Intervention logic\\
\midrule
$c_0r_0$ & Fixed two-specialist panel & Specialist ablation baseline & No challenge or recruitment\\
$c_0r_1$ & Self-triggered specialist recruitment & Specialist ablation & Workers assess missing expertise without external critique\\
$c_1r_0$ & Challenger only & Specialist baseline; combined reference & Critique is injected, but panel stays fixed\\
$c_1r_1$ & Challenger + specialist recruitment & Specialist ablation & Workers reassess team composition after a diagnosed gap\\
$c_1r_0k_1$ & Challenger + knowledge recruitment & Knowledge transfer & Gap-aware in-between retrieval; no specialist recruitment\\
$c_1r_1k_1$ & Both channels, shared context & Naive combination & Recruited specialists receive the evidence brief\\
$c_1r_1k_1^{\mathrm{ind}}$ & Both channels, separated information flow & Coordination redesign & Recruited specialists do not receive the evidence brief directly\\
\bottomrule
\end{tabularx}
\end{table}

\subsection{Agent state and structured outputs}

All roles are instances of the same agent abstraction: a system prompt plus a list of OpenAI-format \texttt{\{role, content\}} messages. State is propagated by appending messages. When the Challenger fires, its critique is appended to each active specialist-role agent's history; recruited agents are initialized with the trajectory available at the recruitment point; and an evidence brief is appended immediately before reconsideration. The Challenger, Knowledge Gate, and related supervisory components use bracketed fields such as \texttt{[Join]: YES/NO}, \texttt{[Knowledge Gap]}, \texttt{[Retrieval Focus]}, and \texttt{[Challenge to Group]}, extracted by regular expressions. Final-answer extraction is constrained to the valid option letters for each dataset, ranging from A through D on most datasets and from A through J on MMLU-Pro and MedXpertQA\@. Generated specialties are free-form role prompts rather than entries from a fixed ontology.

The initial multi-specialist panel is generated freely from the question. The recruiter receives the current panel and is instructed not to duplicate roles. At most two specialists may be added in one turn, at most three may be added over a run, and the total panel is capped at five. Recruited agents receive the discussion available at the point of entry and take part in later turns. The final moderator aggregates the last answers by majority vote. In the independent combined condition, recruited specialists produce targeted auxiliary assessments for the moderator instead of entering the evidence-informed shared discussion.

\subsection{Retrieval configuration}

The Knowledge Gate receives the question, answer options, current reasoning, provisional answer, and, in the Challenger-aware condition, the structured critique. In the upfront condition, retrieval is unconditional and the question text supplies the query. The blind and Challenger-aware in-between gates identify a retrievable gap on most questions; their main difference is whether the Challenger's diagnosis is available when the query is written.

iMedRAG performs two-stage iterative retrieval over PubMed abstracts and medical textbooks. StatPearls is excluded because precomputed embeddings were unavailable. A preliminary test found that adding Wikipedia introduced noise without improving accuracy, so it is also excluded. Retrieval depth is fixed at $k=16$ passages per query; this value was fixed before the reported experiments and was not included in the ablation. Passing the retrieved passages directly to the reasoning agent produced distracting material during preliminary development. The Evidence Curator was therefore introduced to retain factual, non-redundant passages that address the diagnosed gap and to synthesize five to eight evidence points. The live-search validation uses a Claude Haiku model equipped with web search and approximately two to four search calls per question.

\subsection{Evaluation and reproducibility}

All main experiments use temperature zero and seed 42. Each experimental condition is isolated in a separate Python script so that prompt or routing changes cannot leak between conditions. Optuna GridSampler assigns each question to an independent trial; a per-run SQLite database prevents overlap among workers. Concurrency is four questions, while agent turns inside a question remain sequential. Transient rate-limit and connection errors are retried. The primary tables use one pass@1 run, following the benchmark protocol. Section~\ref{app:variance} reports the available three-run GPT-4o-mini analysis. Depending on which gates fire, a turn can add a Challenger call, a Knowledge Gate call, an Evidence Curator call, and recruiter or specialist calls. This amounts to roughly three to five additional model calls per question per turn in the adaptive configurations.

\section{Benchmark context and KAMAC implementation check}
\label{app:benchmark-context}

The main paper relies on controlled comparisons within the proposed pipelines. For context, Table~\ref{tab:baseline-context-app} reproduces the complete GPT-4o-mini baseline table used to characterize the difficulty of the nine-dataset hard subset. The published rows are reported in the first arXiv version of \medbench{} \citep{tang2025medagentsbench}; the KAMAC row is our rerun. The final column is an unweighted mean across the nine datasets. The main paper reports pooled accuracy over 862 questions, so the two aggregate columns should not be compared directly.

\begin{table}[H]
\caption{Benchmark context on the \medbench{} hard subset with GPT-4o-mini. MQA: MedQA; PMQA: PubMedQA; MMCQA: MedMCQA; MB: MedBullets; MP: MMLU-Pro; MEQA: MedExQA; XR and XU: MedXpertQA-R and MedXpertQA-U. Mean is the unweighted mean of the nine dataset accuracies. Published rows come from \medbench{} v1; KAMAC is our rerun.}
\label{tab:baseline-context-app}
\centering
\scriptsize
\begin{adjustbox}{max width=\linewidth}
\begin{tabular}{lrrrrrrrrrr}
\toprule
Method & MQA & PMQA & MMCQA & MB & MMLU & MP & MEQA & XR & XU & Mean\\
\midrule
Zero-shot    & 22.0 & 10.0 & 17.0 & 10.1 & 12.3 & 11.0 &  4.0 &  6.0 &  5.0 & 10.8\\
Few-shot     & 30.0 & 22.0 & 31.0 & 23.6 & 28.8 & 10.0 & 25.0 & 16.0 &  8.0 & 21.6\\
CoT          & 21.0 & 13.0 & 26.0 & 18.0 & 28.8 & 35.0 & 14.0 &  6.0 & 10.0 & 19.1\\
CoT-SC       & 20.0 & 11.0 & 20.0 & 16.9 & 28.8 & 34.0 & 19.0 & 10.0 & 13.0 & 19.2\\
MultiPersona & 29.0 & 13.0 & 21.0 & 15.7 & 26.0 & 36.0 & 17.0 &  7.0 & 12.0 & 19.6\\
Self-Refine  & 32.0 & 12.0 & 24.0 & 15.7 & 27.4 & 31.0 & 16.0 &  7.0 & 12.0 & 19.7\\
MedPrompt    & 29.0 & 14.0 & 30.0 & 13.5 & 20.5 & 34.0 & 18.0 &  6.0 & 13.0 & 19.8\\
MedAgents    & 24.0 & 12.0 & 22.0 & 15.7 & 24.7 &  3.0 & 12.0 &  4.0 &  5.0 & 13.6\\
MDAgents     & 22.0 & 23.0 & 16.0 & 14.6 & 17.8 &  9.0 & 10.0 &  8.0 &  9.0 & 14.4\\
SPO          & 19.0 & 25.0 & 20.0 & 22.5 & 19.2 & 32.0 & 14.0 & 11.0 & 11.0 & 19.3\\
AFlow        & 30.0 & 15.0 & 25.0 & 15.7 & 24.7 & 29.0 &  7.0 &  7.0 &  7.0 & 17.8\\
\midrule
KAMAC        & 21.0 & 17.0 & 24.0 & 15.7 & 24.7 & 30.0 & 11.0 &  7.0 &  8.0 & 17.6\\
\bottomrule
\end{tabular}
\end{adjustbox}
\end{table}

No method is uniformly strong on the hard subset, and the published multi-agent systems do not consistently surpass the prompting baselines. The table is included as benchmark context rather than as a compute-matched comparison with the adaptive pipelines.

Before using KAMAC for the self-assessment diagnostic, we checked the implementation on the full 1,273-question MedQA benchmark. Our run obtained 1,121 correct answers, while the original KAMAC paper reports 1,122 \citep{wu2025knowledge}. We then measured recruitment behavior on the 894-question repository hard set, which includes 32 additional repository questions beyond the primary 862-question benchmark. The resulting recruitment rates are reported in Table~\ref{tab:expert-main}B. The one-question difference reproduces the published full-MedQA result closely before the hard-set analysis.

\section{Full experimental results}
\label{app:fulltables}

The appendix preserves the complete results behind the main-text summaries. Table~\ref{tab:knowledge-full-app} gives the primary knowledge-recruitment ablation; Tables~\ref{tab:expert-full-app},~\ref{tab:expert-medqa-app}, and~\ref{tab:expert-gen-app} give the complementary specialist-recruitment results; and Tables~\ref{tab:combined-4o-app} and~\ref{tab:combined-41-app} report transfer and channel coordination on the multi-specialist backbone.

\subsection{Knowledge-recruitment timing and query signal}

Table~\ref{tab:knowledge-full-app} expands the overall retrieval comparison from Table~\ref{tab:retrieval-main} to all nine datasets. Its no-retrieval column is the Base Clean condition and contains 202 correct answers out of 862.

\begin{table}[H]
\caption{Full knowledge-recruitment ablation on the 862-question hard subset with GPT-4o-mini. The single-generalist backbone is used to isolate retrieval effects. Values are pass@1 accuracy (\%).}
\label{tab:knowledge-full-app}
\centering
\scriptsize
\begin{tabular}{lrrrr}
\toprule
Dataset & No retrieval & Upfront iMedRAG & In-between blind & In-between Challenger-aware\\
\midrule
MedQA        & 34.0 & 31.0 & 34.0 & \textbf{35.0}\\
PubMedQA     & 39.0 & 33.0 & 36.0 & \textbf{43.0}\\
MedMCQA      & 19.0 & 27.0 & 27.0 & \textbf{29.0}\\
MedBullets   & 22.5 & 24.7 & 22.5 & \textbf{27.0}\\
MMLU         & 28.8 & 32.9 & \textbf{34.2} & 32.9\\
MMLU-Pro     & 27.0 & \textbf{33.0} & 29.0 & 32.0\\
MedExQA      & 18.0 & 17.0 & \textbf{21.0} & 18.0\\
MedXpertQA-R & \textbf{13.0} & 10.0 & 7.0 & 11.0\\
MedXpertQA-U & \textbf{11.0} & 7.0 & \textbf{11.0} & \textbf{11.0}\\
\midrule
Overall (862) & 23.4 & 23.7 & 24.4 & \textbf{26.3}\\
\bottomrule
\end{tabular}
\end{table}

\subsection{Complementary specialist recruitment}

Table~\ref{tab:expert-full-app} gives the full GPT-4o-mini $2\times2$ ablation. Table~\ref{tab:expert-medqa-app} gives the complete GPT-4.1-mini MedQA ablation, including the recruitment rate in every condition. Table~\ref{tab:expert-gen-app} reports the additional GPT-4.1-mini subsets.

\begin{table}[H]
\caption{Full $2\times2$ specialist-recruitment ablation on the 862-question hard subset with GPT-4o-mini. Values are pass@1 accuracy (\%).}
\label{tab:expert-full-app}
\centering
\scriptsize
\begin{tabular}{lrrrr}
\toprule
Dataset & Fixed panel $c_0r_0$ & Self-recruit $c_0r_1$ & Challenger $c_1r_0$ & Challenger + recruit $c_1r_1$\\
\midrule
MedQA        & 24.0 & 24.0 & \textbf{28.0} & 25.0\\
PubMedQA     & 15.0 & 14.0 & 19.0 & \textbf{27.0}\\
MedMCQA      & 24.0 & 20.0 & 22.0 & \textbf{26.0}\\
MedBullets   & 18.0 & 16.9 & 23.6 & \textbf{24.7}\\
MMLU         & 20.5 & 21.9 & \textbf{26.0} & 24.7\\
MMLU-Pro     & \textbf{32.0} & 28.0 & \textbf{32.0} & 26.0\\
MedExQA      & 14.0 & 13.0 & 14.0 & \textbf{21.0}\\
MedXpertQA-R &  8.0 &  8.0 & \textbf{9.0} &  7.0\\
MedXpertQA-U &  8.0 &  8.0 & \textbf{11.0} & 10.0\\
\midrule
Overall (862) & 18.1 & 16.9 & 20.3 & \textbf{21.1}\\
\bottomrule
\end{tabular}
\end{table}

\begin{table}[H]
\caption{Focused $2\times2$ specialist-recruitment ablation on 100 MedQA questions with GPT-4.1-mini. Recruitment rate is the fraction of questions on which at least one additional role was added.}
\label{tab:expert-medqa-app}
\centering
\small
\begin{tabular}{lrr}
\toprule
Condition & Accuracy (\%) & Recruitment rate (\%)\\
\midrule
Fixed panel ($c_0r_0$) & 44 & 0\\
Self-triggered recruitment ($c_0r_1$) & 43 & 34\\
Challenger Only ($c_1r_0$) & 46 & 0\\
Challenger + Recruitment ($c_1r_1$) & \textbf{54} & 41\\
\bottomrule
\end{tabular}
\end{table}

\begin{table}[H]
\caption{Cross-dataset specialist-recruitment results with GPT-4.1-mini. Each dataset contains 100 questions. The final column gives the recruitment rate for Challenger + Recruitment.}
\label{tab:expert-gen-app}
\centering
\small
\begin{tabular}{lrrrr}
\toprule
Dataset & Challenger Only & Challenger + Recruitment & Difference & Recruitment rate\\
\midrule
MedQA     & 46 & \textbf{54} & $+8$ & 41\%\\
PubMedQA  & 30 & \textbf{34} & $+4$ & 74\%\\
MedMCQA   & 31 & \textbf{33} & $+2$ & 30\%\\
MMLU-Pro  & 33 & \textbf{40} & $+7$ & 24\%\\
\bottomrule
\end{tabular}
\end{table}

\subsection{Transfer and coordination on the multi-specialist backbone}

Tables~\ref{tab:combined-4o-app} and~\ref{tab:combined-41-app} contain the full knowledge-only and combined-channel comparisons for GPT-4o-mini and GPT-4.1-mini, respectively.

\begin{table}[H]
\caption{Knowledge transfer and combined-channel conditions with GPT-4o-mini on 862 questions. Values are pass@1 accuracy (\%).}
\label{tab:combined-4o-app}
\centering
\scriptsize
\begin{tabular}{lrrrr}
\toprule
Dataset & Challenger only $c_1r_0$ & Knowledge only $c_1r_0k_1$ & Naive both $c_1r_1k_1$ & Independent both $c_1r_1k_1^{\mathrm{ind}}$\\
\midrule
MedQA        & 28.0 & 31.0 & 30.0 & \textbf{35.0}\\
PubMedQA     & 19.0 & \textbf{36.0} & 31.0 & 31.0\\
MedMCQA      & 22.0 & \textbf{25.0} & 19.0 & \textbf{25.0}\\
MedBullets   & 23.6 & 25.8 & 22.5 & \textbf{31.5}\\
MMLU         & 26.0 & \textbf{41.1} & 38.4 & 27.4\\
MMLU-Pro     & 32.0 & \textbf{35.0} & 27.0 & 33.0\\
MedExQA      & 14.0 & \textbf{21.0} & 17.0 & 17.0\\
MedXpertQA-R &  9.0 & \textbf{16.0} & 15.0 &  9.0\\
MedXpertQA-U & \textbf{11.0} & 10.0 & \textbf{11.0} & 8.0\\
\midrule
Overall (862) & 20.3 & \textbf{26.3} & 23.0 & 23.9\\
\bottomrule
\end{tabular}
\end{table}

\begin{table}[H]
\caption{Combined-channel conditions with GPT-4.1-mini on 862 questions. Knowledge-only is the reference for the two combination designs. Values are pass@1 accuracy (\%).}
\label{tab:combined-41-app}
\centering
\scriptsize
\begin{tabular}{lrrr}
\toprule
Dataset & Knowledge only $c_1r_0k_1$ & Naive both $c_1r_1k_1$ & Independent both $c_1r_1k_1^{\mathrm{ind}}$\\
\midrule
MedQA        & 53.0 & 52.0 & \textbf{61.0}\\
PubMedQA     & 33.0 & 31.0 & \textbf{36.0}\\
MedMCQA      & \textbf{32.0} & 28.0 & \textbf{32.0}\\
MedBullets   & \textbf{47.2} & 43.8 & 43.8\\
MMLU         & 41.1 & 41.1 & \textbf{43.8}\\
MMLU-Pro     & \textbf{43.0} & 36.0 & 39.0\\
MedExQA      & 18.0 & \textbf{22.0} & \textbf{22.0}\\
MedXpertQA-R & 30.0 & 26.0 & \textbf{34.0}\\
MedXpertQA-U & \textbf{29.0} & 28.0 & 24.0\\
\midrule
Overall (862) & 36.0 & 33.9 & \textbf{37.0}\\
\bottomrule
\end{tabular}
\end{table}

\section{Transfer to Grok}
\label{app:grok}

All seven conditions were rerun with grok-4-1-fast-non-reasoning without changing prompts or hyperparameters. Table~\ref{tab:grok-app} reports the full results. The same direction appears for Challenger Only over the fixed panel ($24.1\%\rightarrow27.0\%$), specialist recruitment over Challenger Only ($27.0\%\rightarrow32.7\%$), and knowledge recruitment over Challenger Only ($27.0\%\rightarrow35.2\%$). Combined conditions remain below knowledge-only, reinforcing the coordination concern.

\begin{table}[H]
\caption{Accuracy (\%) on the 862-question hard subset using grok-4-1-fast-non-reasoning. MedBullets has 89 questions, MMLU has 73, and all other datasets have 100.}
\label{tab:grok-app}
\centering
\scriptsize
\begin{adjustbox}{max width=\linewidth}
\begin{tabular}{lrrrrrrrrrr}
\toprule
Condition & MedQA & PubMedQA & MedMCQA & MedBullets & MMLU & MMLU-Pro & MedExQA & Xpert-R & Xpert-U & Total\\
\midrule
$c_0r_0$ & 27.0 & 15.0 & 31.0 & 22.5 & 35.6 & 39.0 & 14.0 & 17.0 & 19.0 & 24.1\\
$c_0r_1$ & 25.0 & 12.0 & 35.0 & 25.8 & 31.5 & 38.0 & 16.0 & 15.0 & 19.0 & 23.9\\
$c_1r_0$ & 33.0 & 16.0 & 33.0 & 29.2 & 38.4 & 42.0 & 20.0 & 16.0 & 19.0 & 27.0\\
$c_1r_1$ & 48.0 & 29.0 & 37.0 & 37.1 & 43.8 & 38.0 & 19.0 & 20.0 & 26.0 & 32.7\\
$c_1r_0k_1$ & 46.0 & 34.0 & 34.0 & 44.9 & 39.7 & 44.0 & 19.0 & 28.0 & 29.0 & \textbf{35.2}\\
$c_1r_1k_1$ & 41.0 & 23.0 & 37.0 & 36.0 & 45.2 & 44.0 & 18.0 & 28.0 & 29.0 & 33.1\\
$c_1r_1k_1^{\mathrm{ind}}$ & 38.0 & 23.0 & 38.0 & 30.3 & 52.1 & 40.0 & 21.0 & 26.0 & 29.0 & 32.5\\
\bottomrule
\end{tabular}
\end{adjustbox}
\end{table}

\section{Run-to-run variance on GPT-4o-mini}
\label{app:variance}

We reran all seven GPT-4o-mini conditions three times. Table~\ref{tab:variance-app} reports mean accuracy and sample standard deviation. Total-accuracy standard deviation remains below one percentage point in every condition. The mean results preserve the principal directional comparisons: Challenger Only exceeds the fixed panel, Challenger-aware specialist recruitment exceeds Challenger Only, knowledge-only is strongest, and both combined conditions remain below knowledge-only.

\begin{table}[H]
\caption{Mean accuracy (\%) $\pm$ sample standard deviation across three runs with GPT-4o-mini. Dataset sizes match Table~\ref{tab:grok-app}.}
\label{tab:variance-app}
\centering
\scriptsize
\begin{adjustbox}{max width=\linewidth}
\begin{tabular}{lrrrrrrrrrr}
\toprule
Condition & MedQA & PubMedQA & MedMCQA & MedBullets & MMLU & MMLU-Pro & MedExQA & Xpert-R & Xpert-U & Total\\
\midrule
$c_0r_0$ & $23.0\pm1.0$ & $13.7\pm1.5$ & $22.3\pm2.1$ & $16.9\pm3.0$ & $23.3\pm2.7$ & $32.3\pm0.6$ & $13.3\pm1.2$ & $6.0\pm2.0$ & $8.7\pm1.2$ & $17.6\pm0.5$\\
$c_0r_1$ & $25.3\pm1.2$ & $13.3\pm2.1$ & $23.0\pm2.6$ & $16.5\pm1.7$ & $19.6\pm2.1$ & $30.3\pm2.1$ & $12.7\pm0.6$ & $6.3\pm1.5$ & $10.0\pm2.0$ & $17.4\pm0.5$\\
$c_1r_0$ & $28.7\pm1.2$ & $20.0\pm2.6$ & $21.7\pm1.5$ & $22.8\pm1.3$ & $25.6\pm2.1$ & $31.7\pm1.5$ & $17.3\pm3.1$ & $8.0\pm1.0$ & $10.0\pm1.7$ & $20.5\pm0.3$\\
$c_1r_1$ & $25.7\pm1.2$ & $27.7\pm1.2$ & $23.7\pm2.5$ & $24.7\pm0.0$ & $29.7\pm4.4$ & $30.7\pm4.5$ & $19.3\pm2.1$ & $6.3\pm3.1$ & $9.3\pm2.1$ & $21.6\pm0.6$\\
$c_1r_0k_1$ & $31.3\pm0.6$ & $38.3\pm2.1$ & $25.7\pm3.1$ & $27.3\pm2.6$ & $37.4\pm4.2$ & $34.0\pm1.7$ & $21.3\pm1.5$ & $19.0\pm3.0$ & $9.3\pm1.2$ & $26.8\pm0.4$\\
$c_1r_1k_1$ & $31.3\pm2.3$ & $34.3\pm2.9$ & $21.3\pm2.1$ & $25.1\pm2.3$ & $32.9\pm5.5$ & $27.0\pm2.0$ & $18.0\pm1.0$ & $15.3\pm0.6$ & $10.7\pm0.6$ & $23.7\pm0.7$\\
$c_1r_1k_1^{\mathrm{ind}}$ & $33.3\pm2.1$ & $34.3\pm3.1$ & $24.0\pm1.7$ & $27.0\pm3.9$ & $29.7\pm2.9$ & $30.7\pm2.1$ & $20.3\pm2.9$ & $11.7\pm2.3$ & $6.7\pm2.3$ & $24.0\pm0.1$\\
\bottomrule
\end{tabular}
\end{adjustbox}
\end{table}

\section{Qualitative medical case studies}
\label{app:cases}

The three cases cover an answer-relevant evidence gap, retrieval anchoring, and a missing specialist perspective. Knowledge recruitment is examined first, followed by the complementary specialist-recruitment example.

\subsection{Challenger-aware knowledge retrieval: MedQA-44}

\paragraph{Question and initial panel.}
MedQA question medqa-44 describes a woman with abnormal uterine bleeding and pelvic symptoms and asks which physical-examination finding is most likely. The correct option is (A), a globular uterus approximately the size of a ten-week pregnancy, consistent with adenomyosis. The initial panel contains a gynecologist and a radiologist. The gynecologist favors adenomyosis and supports (A), while the radiologist favors fibroids and supports (C). Both positions are medically plausible, but the discussion does not identify a fact that separates the expected examination findings.

\paragraph{Challenger and retrieval.}
The Challenger repeatedly raises concern about endometrial pathology given the patient's age, obesity, diabetes, and irregular bleeding. A representative part of its critique is:
\begin{quote}\small
``The current reasoning overlooks the critical need to exclude endometrial pathology, including hyperplasia or malignancy, especially given her age, obesity, diabetes, and irregular bleeding with intermenstrual spotting.''
\end{quote}
The concern is relevant to the presentation, but it does not resolve the anatomical distinction between adenomyosis and fibroids. Challenger Only therefore ends on (C). With knowledge recruitment enabled, the Knowledge Gate formulates a more specific request: physical-examination findings that distinguish adenomyosis from fibroids. The resulting brief states that adenomyosis typically produces a moderately enlarged, globular, and often tender uterus approximately the size of a ten-week pregnancy, whereas fibroids more often produce a larger, irregularly contoured or nodular uterus.

\paragraph{Outcome.}
The evidence supplies the missing comparison, and the panel converges on (A). Table~\ref{tab:case-44-app} summarizes the trace. This case separates the two supervisory functions: the Challenger reopens the reasoning, while the Knowledge Gate identifies the retrievable question that directly distinguishes the answer options.

\begin{table}[H]
\caption{Trace summary for MedQA-44.}
\label{tab:case-44-app}
\centering
\small
\begin{tabularx}{\linewidth}{>{\raggedright\arraybackslash}p{30mm}Y}
\toprule
Component & Trace content\\
\midrule
Initial panel & Gynecologist favors adenomyosis (A); radiologist favors fibroids (C).\\
Challenger & Raises malignancy risk but does not settle the physical-examination distinction.\\
Knowledge Gate focus & ``Physical examination findings that distinguish adenomyosis from fibroids.''\\
Evidence effect & A globular uterus approximately the size of a ten-week pregnancy supports adenomyosis; a larger irregular contour supports fibroids.\\
Outcome & Challenger Only: (C), incorrect. In-between retrieval: (A), correct.\\
\bottomrule
\end{tabularx}
\end{table}

\subsection{Retrieval timing and anchoring: MedQA-477}

\paragraph{Question and initial interpretation.}
MedQA question medqa-477 describes a man who develops sensory loss over the lateral forearm after heavy lifting. Thumb sensation is intact and cervical pathology has been ruled out. The question asks which motor weakness is most likely. The correct option is (A), weakness in elbow flexion caused by musculocutaneous nerve injury. Both runs initially consider a radial-nerve explanation.

\paragraph{Upfront retrieval.}
In the upfront condition, iMedRAG is called before the first answer and returns material centered on the radial nerve. The evidence is accurate on its own terms:
\begin{quote}\small
``The radial nerve innervates the extensor muscles of the forearm, which are responsible for wrist and finger extension. Injury to the radial nerve can lead to wrist drop.''
\end{quote}
The Challenger later notes that lateral-forearm sensory loss could instead implicate the musculocutaneous or lateral antebrachial cutaneous nerve, but the radial framing remains dominant. The moderator finishes on (D), weakness in wrist extension.

\paragraph{In-between retrieval.}
In the in-between condition, the agents produce an initial interpretation before any evidence is injected. The Challenger focuses on the sensory distribution and questions the radial-nerve account. The Knowledge Gate records the gap as uncertainty about the nerve responsible for lateral-forearm sensation and asks for evidence on musculocutaneous involvement. The retrieved brief links the musculocutaneous nerve to the biceps and brachialis, elbow flexion, and lateral-forearm sensation. The moderator revises to (A).

\paragraph{Outcome.}
Table~\ref{tab:case-477-app} compares the two traces. The contrast is not between false and true evidence. Both retrievals return plausible medical facts. The difference is whether the query targets the anatomical distinction that has become disputed during reasoning.

\begin{table}[H]
\caption{Trace summary for MedQA-477.}
\label{tab:case-477-app}
\centering
\small
\begin{tabularx}{\linewidth}{>{\raggedright\arraybackslash}p{30mm}YY}
\toprule
Component & Upfront retrieval & In-between retrieval\\
\midrule
Retrieved target & Radial nerve and wrist extension & Musculocutaneous nerve and lateral-forearm innervation\\
Challenger role & Mentions alternatives after the radial frame is established & Exposes the sensory-distribution inconsistency before evidence is retrieved\\
Final answer & (D), incorrect & (A), correct\\
\bottomrule
\end{tabularx}
\end{table}

\subsection{Complementary Challenger-aware specialist recruitment: MedQA-34}

\paragraph{Question and initial panel.}
MedQA question medqa-34 describes a 27-year-old man brought in unresponsive with glucose 589 mg/dL, potassium 6.3 mEq/L, and bicarbonate 17 mEq/L. After initial treatment, glucose falls to 90 mg/dL and potassium to 4.3 mEq/L, but bicarbonate remains at 19 mEq/L. The correct option is (A), continuing insulin, glucose supplementation, potassium management, and intravenous fluids. The initial team contains an emergency physician and an endocrinologist. One agent favors supportive therapy and close monitoring (D), while the other supports continued insulin-based management (A).

\paragraph{Challenger diagnosis.}
The Challenger focuses on the persistent bicarbonate abnormality and the interaction between insulin and electrolyte management. Its critique states that the panel is underestimating incomplete resolution of the acidosis and should reason jointly about insulin, glucose, and potassium. Without recruitment, the discussion does not convert this point into a common decision, and the moderator selects (D).

\paragraph{Recruited perspective.}
With recruitment enabled, the system adds a clinical pharmacologist with expertise in pharmacokinetics and electrolyte management. The recruited role emphasizes that insulin must continue to suppress ketogenesis, that glucose supplementation may be needed while insulin continues, and that potassium must be monitored because insulin shifts it intracellularly. It concludes that supportive monitoring alone is insufficient while the metabolic disturbance remains unresolved.

\paragraph{Outcome.}
The panel converges on (A). Table~\ref{tab:case-34-app} summarizes the intervention. The initial panel already contains two broadly appropriate specialties; the recruited role is useful because it reorganizes the discussion around treatment management and the interaction among the remaining abnormalities.

\begin{table}[H]
\caption{Trace summary for MedQA-34.}
\label{tab:case-34-app}
\centering
\small
\begin{tabularx}{\linewidth}{>{\raggedright\arraybackslash}p{30mm}Y}
\toprule
Component & Trace content\\
\midrule
Initial panel & Emergency physician and endocrinologist disagree between supportive monitoring (D) and continued insulin-based management (A).\\
Challenger diagnosis & Persistent bicarbonate of 19 mEq/L indicates incomplete resolution; the panel has not jointly reasoned about insulin, glucose, and potassium.\\
Recruited specialist & Clinical pharmacologist with electrolyte-management expertise argues that supportive monitoring alone is insufficient.\\
Outcome & Challenger Only: (D), incorrect. Challenger + Recruitment: (A), correct.\\
\bottomrule
\end{tabularx}
\end{table}

\section{Cross-case medical discussion}
\label{app:medical-discussion}

\paragraph{Useful retrieval queries are local to the disputed distinction.}
The MedQA-44 and MedQA-477 cases show that broad topical relevance is not enough. The query must distinguish the options currently under consideration. In one case, that distinction is uterine contour for adenomyosis versus fibroids. In the other, it is lateral-forearm innervation for musculocutaneous versus radial nerve injury.

\paragraph{The Challenger and Knowledge Gate have different jobs.}
The Challenger reopens the reasoning and identifies a weakness. The Knowledge Gate decides whether that weakness can be answered with evidence and writes the query. MedQA-44 illustrates why the separation matters: the Challenger raises a related malignancy concern, while the Knowledge Gate retrieves the anatomical comparison that resolves the question.

\paragraph{Specialist recruitment changes the available perspective.}
In MedQA-34, the initial panel already includes emergency medicine and endocrinology roles. Recruitment is useful because the new role emphasizes treatment management across insulin, glucose, potassium, and fluids. The intervention adds a different way of organizing the existing facts rather than simply repeating the initial discussion.

\paragraph{Channel selection and aggregation remain important.}
The cases suggest a natural routing policy. Factual distinctions can be directed to retrieval, missing perspectives to specialist recruitment, and unresolved reasoning conflicts to further critique. When several channels are activated together, the moderator must account for shared evidence and correlated arguments rather than treating every vote as independent.

\end{document}